\documentclass[journal]{IEEEtran}
\usepackage[table]{xcolor}
\usepackage{colortbl}
\usepackage{amsmath,amsfonts}
\pdfoutput=1
\usepackage{algorithm}
\usepackage{array}
\usepackage[caption=false,font=normalsize,labelfont=sf,textfont=sf]{subfig}
\usepackage{textcomp}
\usepackage{stfloats}
\usepackage{url}
\usepackage{verbatim}
\usepackage{graphicx}
\usepackage{subfig}
\usepackage{cite}
\usepackage{float}
\usepackage[export]{adjustbox} 
\usepackage{siunitx}
\usepackage{multirow}
\usepackage{threeparttable}
\usepackage{color}
\usepackage{amsmath, bm}
\usepackage{bbding}
\usepackage[backref]{hyperref}
\usepackage{pifont}
\usepackage[all]{hypcap}
\usepackage{makecell}
\usepackage{lineno}
\usepackage{arydshln}
\usepackage{amsmath} 
\usepackage{amssymb}
\usepackage{xcolor}
\usepackage{mathtools}
\usepackage{multirow}

\usepackage{algorithm}
\usepackage{booktabs}
\usepackage{cuted}   
\usepackage{cleveref}
\usepackage{subcaption}
\usepackage{adjustbox}
\usepackage{algorithm}
\usepackage{algpseudocode}
\usepackage{lipsum}
\usepackage{array}
\newcolumntype{M}[1]{>{\centering\arraybackslash}m{#1}}  % 自定义垂直居中列

\usepackage{array} % Required for m{<width>} column type
\usepackage{booktabs} % For professional table formatting
\usepackage{amsmath} % For math symbols
\usepackage{multirow} % For multi-row cells, if needed

\begin{document}
\title{GRASPTrack: Geometry-Reasoned Association via Segmentation and Projection
for Multi-Object Tracking}

\author{\IEEEauthorblockN{Xudong Han\IEEEauthorrefmark{1},
                        Pengcheng Fang\IEEEauthorrefmark{1},
                         Yueying Tian,
                        Jianhui Yu,
                        Xiaohao Cai,
                        Daniel Roggen, 
                         Philip Birch \IEEEauthorrefmark{2},
                        }
                        
% <-this % stops a space
\thanks{\IEEEauthorrefmark{1} Equal Contribution.
\\
\IEEEauthorrefmark{2} Corresponding author.
}}

\markboth{Journal of \LaTeX\ Class Files,~Vol.~14, No.~8, August~2021}%
{Shell \MakeLowercase{\textit{et al.}}: A Sample Article Using IEEEtran.cls for IEEE Journals}

\maketitle

\begin{abstract}

Multi-object tracking (MOT) in monocular videos is fundamentally challenged by occlusions and depth ambiguity, issues that conventional tracking-by-detection (TBD) methods struggle to resolve owing to a lack of geometric awareness. To address these limitations, we introduce GRASPTrack, a novel depth-aware MOT framework that integrates monocular depth estimation and instance segmentation into a standard TBD pipeline to generate high-fidelity 3D point clouds from 2D detections, thereby enabling explicit 3D geometric reasoning. These 3D point clouds are then voxelized to enable a precise and robust Voxel-Based 3D Intersection-over-Union (IoU) for spatial association. To further enhance tracking robustness, our approach incorporates Depth-aware Adaptive Noise Compensation, which dynamically adjusts the Kalman filter process noise based on occlusion severity for more reliable state estimation. Additionally, we propose a Depth-enhanced Observation-Centric Momentum, which extends the motion direction consistency from the image plane into 3D space to improve motion-based association cues, particularly for objects with complex trajectories. Extensive experiments on the MOT17, MOT20, and DanceTrack benchmarks demonstrate that our method achieves competitive performance, significantly improving tracking robustness in complex scenes with frequent occlusions and intricate motion patterns.

\end{abstract}



% \input{sections/intro.tex}
% \input{sections/background_related_work.tex}
% % \input{sections/problem_setup.tex}
% \input{sections/method.tex}
% \input{sections/experiments.tex}
% \input{sections/conclusion_ackno.tex}

\section{Introduction}
\label{sec:intro}
Multi-object tracking (MOT) is a critical task in computer vision with many applications, such as autonomous driving \cite{sun2020scalability}, robotic navigation \cite{yuan2022glamr}, and sports analytics \cite{torres2022tracking}. Most MOT methods typically follow the tracking-by-detection (TBD) paradigm, where objects are detected independently in each frame and associated across frames based on motion and appearance cues. These MOT methods typically rely on 2D bounding box detection and frame-wise association through metrics such as the Intersection-over-Union (IoU). Despite their efficiency, these approaches inherently lack geometric awareness, making them vulnerable to object interactions, depth ambiguity, and occlusions.

Current MOT methods face several challenges in real-world scenarios. One critical problem is the occlusion. When multiple objects at different depths overlap in the 2D image plane, even short-term partial occlusions can result in heavy overlap, leading to identity switches that IoU-based matching struggles to resolve. Another significant challenge is accurately modelling motion. For instance, objects moving along the optical axis of the camera may undergo substantial 3D motion with minimal 2D positional changes, leading to erroneous velocity estimates and association failures. To mitigate such issues, several existing works \cite{liu2025sparsetrack, limanta2024camot} attempt to infer pseudo-depth from 2D cues. However, these methods rely on strong scene assumptions and typically produce imprecise depth estimations. In addition, other methods \cite{dendorfer2022quo,khurana2021detecting} utilize a monocular depth estimation model to obtain depth maps, but typically extract 3D features from the entire 2D bounding box. This process introduces significant noise from the background and even occluding objects, degrading the quality of the object’s 3D representation.

\begin{figure}[!t]
    \centering
    \includegraphics[width=1.0\linewidth]{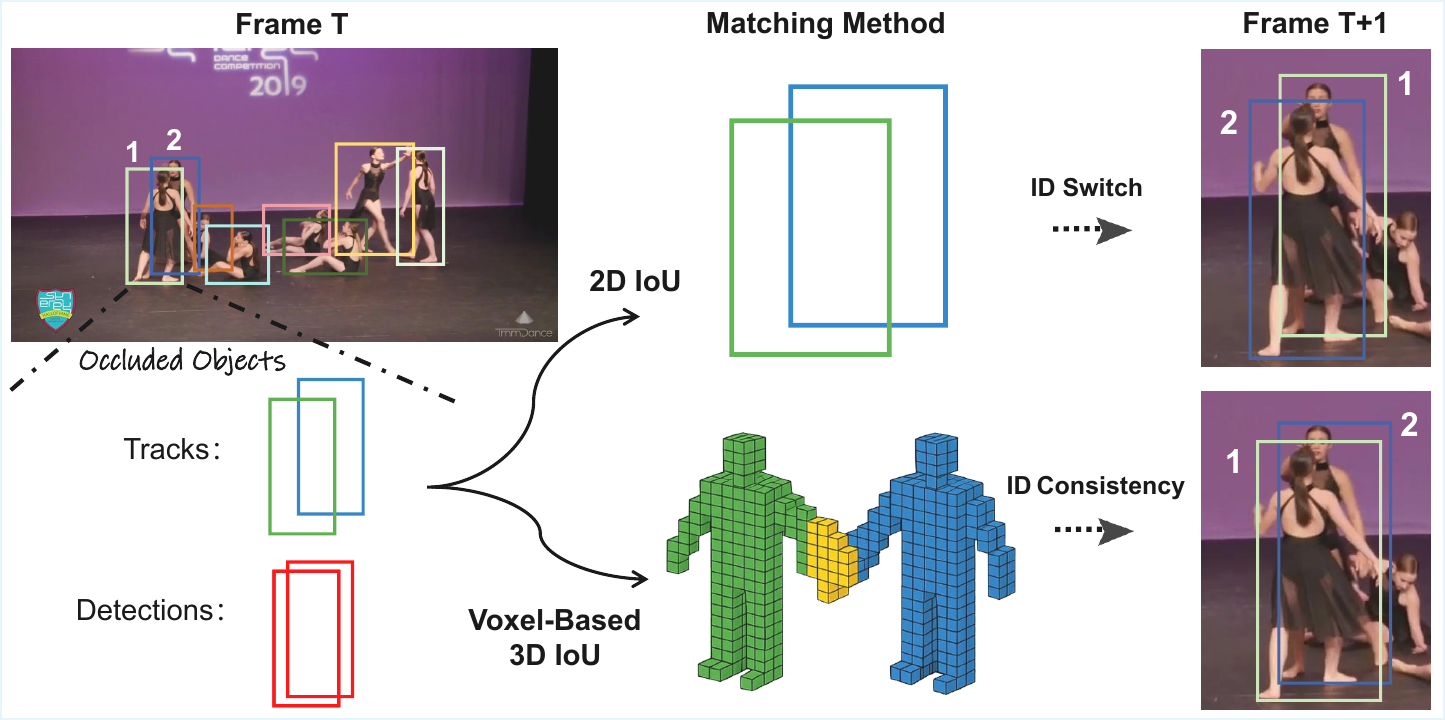}
    \caption{An illustration of associating occluded detections in crowded scenes. In the presence of heavy occlusion, conventional 2D IoU-based matching can lead to ID switches owing to spatial ambiguity between overlapping objects. To address this, we propose a Voxel-Based IoU metric that operates in 3D space, enabling more accurate association by capturing fine-grained volumetric overlap and handling partial occlusions with improved spatial reasoning.}
    \label{fig:voxel}
\end{figure}

To address these limitations, this study proposes a depth-aware MOT framework that explicitly incorporates geometric reasoning into the tracking pipeline, called GRASPTrack. Our approach leverages advanced models in monocular depth estimation and segmentation to enrich scene understanding from a single image. Specifically, we use a segmentation model to generate a precise instance mask for each object. This mask guides the creation of a clean, high-fidelity 3D point cloud from the dense depth map produced by a monocular depth estimation model. To enhance spatial matching, these point clouds are transformed into voxel representations, enabling a Voxel-Based 3D IoU for robust association and better reflecting their true spatial extent.

Additionally, we enhance motion modeling in the presence of occlusions. Traditional Kalman Filters \cite{kalman1960contributions} rely on fixed process noise assumptions, which fail to adapt to the increased uncertainty introduced by occlusions.
We propose a Depth-aware Adaptive Noise Compensation (DANC) method that dynamically adjusts the process noise covariance in the Kalman filter based on the severity of occlusion, ensuring more conservative and reliable state updates under uncertainty. Furthermore, the Observation-Centric Momentum (OCM) introduced in OC-SORT \cite{cao2023observation} leverages motion direction consistency to improve association robustness. We introduce a Depth-enhanced Observation-Centric Momentum (DOCM) to extend motion direction consistency modeling from 2D to 3D space. By calculating the motion direction consistency using full 3D state vectors, our method provides a more robust motion cue, leading to more reliable data association. We evaluate our method on several challenging datasets, such as MOT17 \cite{milan2016mot16}, MOT20 \cite{dendorfer2020mot20} and DanceTrack \cite{sun2022dancetrack}. Experimental results show that our method achieves highly competitive performance among tracking-by-detection methods.

The main contributions of this study are as follows:
\begin{itemize}
    \item We propose GRASPTrack, a novel depth-aware MOT framework that integrates geometric reasoning into the tracking pipeline, significantly enhancing robustness under occlusion. We leverage monocular depth estimation and segmentation masks to reconstruct high-fidelity 3D point clouds from 2D detection. These are voxelized to enable Voxel-Based 3D IoU for object association, while mask-guided refinement effectively suppresses background and occluder noise.
    \item We introduce DANC, a dynamic Kalman filter process noise adjustment mechanism that accounts for occlusion severity. In addition, we extend the Kalman filter state vector using depth information to enable accurate spatial state estimation in 3D space.
    \item We propose DOCM to extend the motion direction consistency in 3D space, improving motion-based association under complex scenarios.
    \item Extensive experimental results and comparison are conducted on challenging benchmark datasets.
\end{itemize}

\begin{figure*}[htbp]
    \centering
    \includegraphics[width=1.0\linewidth]{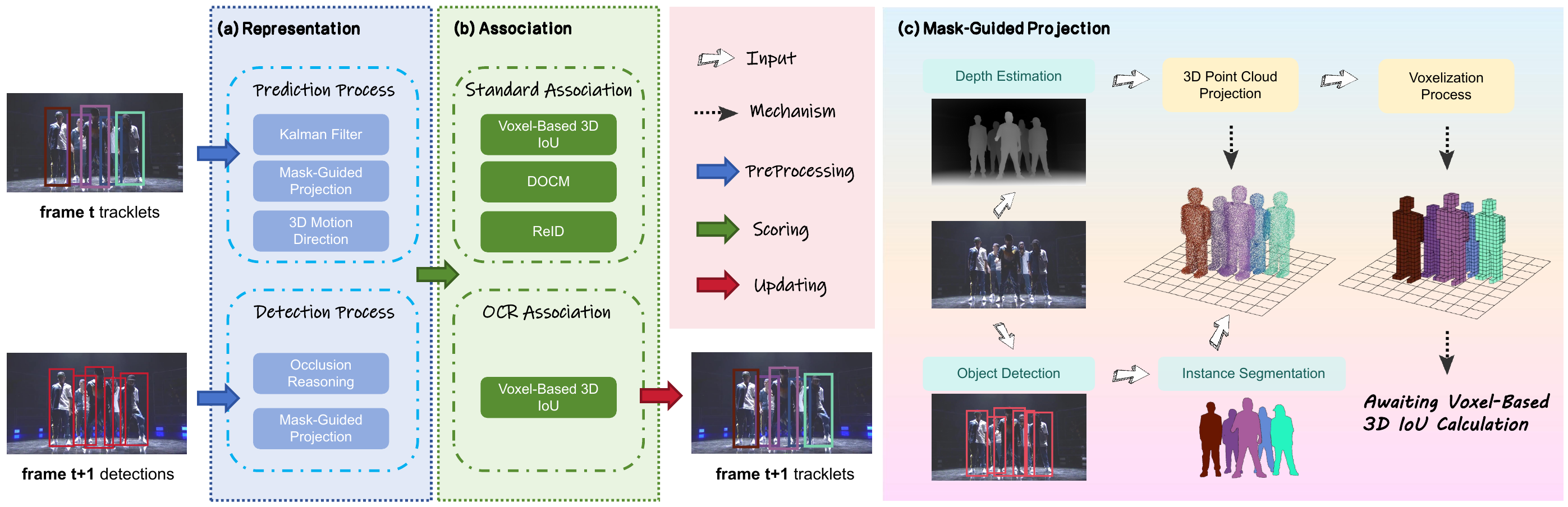}
    \caption{The pipeline of the proposed GRASPTrack. \textbf{(a)} Representation: We estimate the probable location of each tracked object in the current frame by leveraging a Kalman filter. To reconstruct the 3D geometry of an object, we apply a mask-guided projection that combines depth estimation and instance segmentation cues. To facilitate more accurate and efficient association, the resulting point cloud is voxelized to enable voxel-based 3D reasoning. For each detection, we first assess its occlusion state and accordingly adapt the Kalman filter's noise covariance in subsequent frames; \textbf{(b)} Association: We compute the Voxel-Based 3D IoU and DOCM (Depth-enhanced Observation-Centric Momentum) to capture geometric similarity, while appearance similarity is measured using a ReID model. In the OCR association stage, only the Voxel-Based 3D IoU is employed; and \textbf{(c)} Process of our Mask-Guided Projection.
    }
    \label{fig:example}
\end{figure*}

\section{Background and Related Work}
\label{sec:background_related_work}

\subsection{Tracking by Detection}

Many current multi-object tracking methods follow the TBD paradigm \cite{bewley2016simple, zhang2022bytetrack, cao2023observation, han2025ettrack}. These methods use a detector to detect objects in each frame and associate them across various frames. Early TBD methods, such as SORT \cite{bewley2016simple}, relied on the Kalman Filter for motion prediction and the IoU between predicted and detected bounding boxes for association. DeepSORT \cite{wojke2017simple} introduced a ReID-based appearance similarity in the cost matrix to enhance robustness  and handle longer-term occlusions where the IoU would fail. ByteTrack \cite{zhang2022bytetrack} introduced a simple and effective heuristic for associating low-confidence detections separately to recover objects during occlusion. OCSORT \cite{cao2023observation} enhanced the robustness of handling occlusions by improving the linear motion assumption in the Kalman filter. Deep OC-SORT \cite{maggiolino2023deep} integrated appearance features and camera motion compensation. UCMCTrack \cite{yi2024ucmctrack} proposed a method that handles camera motion in object tracking by replacing the standard IoU metric with a Mapped Mahalanobis Distance on the ground plane. TBD methods have shown that the combination of a strong detector with a simple association strategy can yield competitive tracking performance. Therefore, we chose to follow the TBD paradigm in this study.

\subsection{Depth Information in MOT}
Adding depth information as a form of spatial context is a key strategy for making multi-object tracking more robust, particularly in crowded scenes. In the domain of 3D MOT, trackers such as AB3DMOT \cite{weng2020ab3dmot} and CenterPoint \cite{yin2021center} leverage explicit 3D sensors, such as LiDAR, to track objects in true 3D space. However, these approaches depend on specialized and costly hardware, which restricts their widespread application. This has motivated the development of methods that can infer 3D information from a more accessible single 2D image, which implicitly contains depth cues through perspective projection. Approaches using a single camera have largely followed two directions. The first uses pseudo-depth heuristics to infer a relative depth order from an object's position in the 2D frame. SparseTrack \cite{liu2025sparsetrack} leveraged pseudo-depth to separate objects along the depth axis and divided the detected objects into multiple sparse subsets at different depths. CAMOT \cite{limanta2024camot}  incorporated a pseudo-depth state directly into its Kalman filter. The second direction involves the use of a monocular depth estimation model to generate a depth map. QuoVadis \cite{dendorfer2022quo} used these maps to create a bird's-eye view (BEV) representation for forecasting. However, these prior studies are limited because they either relied on coarse geometric heuristics or used depth information merely as an auxiliary cue to improve tracking performance. In this study, we propose a more robust and holistic integration of 3D geometric reasoning by integrating more precise depth information to enhance the robustness of the tracker in complex and occluded scenes.

\section{Method}
\label{sec:method}
% In this paper, GRASPTrack follows the tracking-by-detection paradigm, with the overall pipeline illustrated in Figure \ref{fig:example}. In this section, we present a depth-aware MOT framework that substantially improves conventional 2D tracking pipelines by explicitly incorporating 3D geometric cues. Our approach introduces three core innovations: (1) Mask-guided voxel-based matching strategy, which reconstructs accurate 3D point clouds using monocular depth estimation and instance segmentation, enabling Voxel-Based 3D IoU for robust association; (2) Depth-Aware Adaptive Noise Compensation that incorporates segmentation-based depth estimation and occlusion-driven noise adjustment for robust 3D state prediction; and (3) a Depth-Enhanced Observation-Centric Momentum (DOCM) module that models 3D motion direction consistency to improve motion-based association. These three components form a unified framework that enhances tracking accuracy in challenging scenarios involving occlusions, depth ambiguity, and complex scene dynamics.

GRASPTrack enhances the TBD paradigm with a depth-aware framework composed of three main components. We first introduce a Depth-Aware Voxelization and 3D IoU Computation module, which converts segmented depth maps into voxel grids for geometric matching. This is followed by a DANC module that incorporates depth cues into state prediction. Finally, a DOCM module models motion consistency in 3D space. All components are coherently designed around depth, forming a fully integrated framework for depth-aware multi-object tracking.

\subsection{Depth-Aware Voxelization and 3D IoU}
%\subsection{Voxel-Based 3D IoU}
As illustrated in Figure \ref{fig:example}(c), GRASPTrack recovers accurate 3D spatial representations of objects from monocular RGB images. The monocular image is fed into two foundational models in parallel: Depth Anything v2 \cite{yang2024depth}, which performs high-quality depth estimation with enhanced cross-scene generalization and improved reconstruction of fine-grained depth details, and EfficientTAM \cite{xiong2024efficient}, which generates segmentation masks for the objects using box prompts. For each input frame $I_t \in \mathbb{R}^{H \times W \times 3}$, we estimate a dense depth map $D_t \in \mathbb{R}^{H \times W}$ using a monocular depth estimation network as follows:
\begin{equation}
D_t = f_{\text{depth}}(I_t),
\end{equation}
where $f_{\text{depth}}(\cdot)$ denotes the depth estimation model.

Given a set of bounding boxes $\mathcal{B}_t = \{b_i^t\}_{i=1}^{N_t}$ from both the detector and tracker, with each $b_i^t = [x_1^i, y_1^i, x_2^i, y_2^i, s_i^t]$, we obtain the corresponding binary segmentation masks using EfficientTAM, i.e.,
\begin{equation}
M_i^t = f_{\text{seg}}(I_t, b_i^t), \quad M_i^t \in \{0,1\}^{H \times W},
\end{equation}
where $M_i^t(u,v) = 1$ indicates that pixel $(u,v)$ belongs to object $i$ at time $t$.

\subsubsection{Mask-Guided Projection}
Using the estimated depth map $D_t$ and mask $M_i^t$, we reconstruct a per-object 3D point cloud by projecting pixels within the mask region into camera coordinates using the standard camera model. For each pixel within the segmentation mask, the 3D coordinates are computed using the standard pinhole camera projection equations:
\begin{equation}
\begin{aligned}
Z &= D_t(u,v), \ 
X = \frac{(u - c_x) Z}{f_x}, \ 
Y = \frac{(v - c_y) Z}{f_y},
\end{aligned}
\end{equation}
where $(u,v)$ are pixel coordinates of the projection plane, $Z$ is the depth value, $(c_x, c_y)$ is the center point of the box corresponding to the object, and $(f_x, f_y)$ are the focal lengths in the $x$ and $y$ directions, respectively.
For each object $i$ at time $t$, we construct a 3D point cloud by collecting all valid projected points within its segmentation mask, i.e.,
\begin{equation}
\mathcal{P}_i^t = \{ \mathbf{p} = [X, Y, Z]^\top \mid M_i^t(u,v) = 1, D_t(u,v) > 0 \},
\end{equation}
where $M_i^t(u,v) = 1$ indicates pixels within the object's segmentation mask, and $D_t(u,v) > 0$ ensures valid depth values. This formulation ensures that only valid depth values within the precise segmentation boundary of the object are considered, providing a more accurate 3D representation than using entire bounding boxes.
This mask-guided projection eliminates the background and occluder pixels, ensuring that only valid object regions contribute to the 3D geometry. 

\subsubsection{Voxelization Process} %Voxel-Based 3D IoU

While 3D point clouds $\mathcal{P}_i^t$ offer fine-grained geometric details, traditional 3D IoU computations typically rely on fitting coarse 3D bounding boxes, which fail to capture the true object shape \cite{shin2024spherical}. To better preserve geometric fidelity while enabling efficient pairwise comparison, we adopt a voxel-based representation that discretizes each $\mathcal{P}_i^t$ into a binary occupancy grid. This allows us to compute the 3D Intersection-over-Union (IoU) directly on the volumetric shape, yielding a more accurate and robust similarity metric. Unlike the voxelization adopted in detection frameworks \cite{zhou2018voxelnet}, which is used solely for feature extraction before regressing a bounding box, our voxel grid is used exclusively during evaluation. Each voxel stores a single binary occupancy value and does not participate in network training or inference.

To ensure consistent voxelization across different frames and object pairs, we establish a unified 3D coordinate system. Given two sets of 3D point clouds $\mathcal{P}_i^{\text{det}}$ and $\mathcal{P}_j^{\text{trk}}$ representing  detections and tracks respectively, we compute the overall spatial bounds:
\begin{align}
\mathbf{p}_{\min} &= \min(\min_{\mathbf{p} \in \mathcal{P}_i^{\text{det}}} \mathbf{p}, \min_{\mathbf{p} \in \mathcal{P}_j^{\text{trk}}} \mathbf{p}), \\
\mathbf{p}_{\max} &= \max(\max_{\mathbf{p} \in \mathcal{P}_i^{\text{det}}} \mathbf{p}, \max_{\mathbf{p} \in \mathcal{P}_j^{\text{trk}}} \mathbf{p}),
\end{align}
where $\mathbf{p}_{\min}, \mathbf{p}_{\max} \in \mathbb{R}^3$ define the global 3D bounding volume that encompasses both point clouds. This way ensures that all the point clouds share the same voxel coordinate.

We discretize the continuous 3D space into a regular voxel grid using a voxel size parameter $\delta_v$, which determines the spatial resolution of the discretization. The voxel size $\delta_v$ controls the fundamental trade-off between computational efficiency and spatial precision, smaller values provide finer granularity but increase memory usage and computation time. In our implementation, we set $\delta_v = 0.4$ to balance the accuracy and efficiency for typical object scales in multi-object tracking scenarios.

For each point cloud $\mathcal{P}$, we transform the 3D coordinates into discrete voxel indices as follows:
\begin{equation}
\mathbf{v}(\mathbf{p}) = \left\lfloor \frac{\mathbf{p} - \mathbf{p}_{\min}}{\delta_v} \right\rfloor,
\end{equation}
where $\mathbf{p} = [x, y, z]^\top$ is a 3D point, and $\mathbf{v}(\mathbf{p})$ represents the corresponding voxel index. To ensure valid indices, we apply boundary constraints to keep all indices within the computed grid dimensions. Since multiple points may map to the same voxel, we perform de-duplication by retaining only unique voxel indices.

We create a sparse binary occupancy grid $\mathcal{V} \in \{0,1\}^{N_x \times N_y \times N_z}$ where each voxel is marked as occupied if it contains at least one point from the object. This sparse representation is crucial for computational efficiency because typical object point clouds occupy only a small fraction of the total voxel space. The resulting occupancy grid provides a discretized volumetric representation that captures the essential 3D structure of each object while enabling efficient intersection and union operations for the IoU computation. Building upon this sparse volumetric encoding, we next describe how 3D IoU is efficiently computed between voxelized objects.

\subsubsection{Voxel-Based 3D IoU Computation}
Given two voxelized occupancy grids $\mathcal{V}_i$ and $\mathcal{V}_j$ representing objects $i$ and $j$ respectively, we compute the 3D IoU between objects, as illustrated in Figure \ref{fig:voxel}, following the standard intersection-over-union formulation adapted to voxelized volumes, i.e.,
\begin{equation}
\text{IoU}_{3D}(\mathcal{V}_i, \mathcal{V}_j) = \frac{|\mathcal{V}_i \cap \mathcal{V}_j|}{|\mathcal{V}_i \cup \mathcal{V}_j|}.
\end{equation}
The intersection $|\mathcal{V}_i \cap \mathcal{V}_j|$ counts the number of voxels occupied in both grids, which is computed through element-wise logical {\tt AND} operations across all voxel positions. Similarly, the union $|\mathcal{V}_i \cup \mathcal{V}_j|$ counts the voxels occupied in either grid, obtained through element-wise logical {\tt OR} operations. This voxel-based IoU computation provides several advantages over the traditional 2D IoU. First, it captures precise volumetric overlap rather than only projected area overlap, making it robust to changes in the viewpoint and camera motion. Second, it naturally handles complex object shapes and partial occlusions by considering the 3D spatial occupancy. 

\begin{table*}[t]
\centering
\footnotesize
\setlength{\tabcolsep}{4.5pt}  % 控制列间距，更小值更紧凑
\renewcommand{\arraystretch}{1.1} % 控制行高，更小值更紧凑
\begin{tabular}{l|cccc|cccc}
\hline
\multirow{2}{*}{Tracker} & \multicolumn{4}{c|}{\textbf{MOT17}} & \multicolumn{4}{c}{\textbf{MOT20}} \\
& HOTA$\uparrow$ & IDF1$\uparrow$ & MOTA$\uparrow$ & AssA$\uparrow$ & HOTA$\uparrow$ & IDF1$\uparrow$ & MOTA$\uparrow$ & AssA$\uparrow$ \\
\hline
\multicolumn{9}{l}{\textcolor{black}{\textbf{Motion:}}} \\
ByteTrack \cite{zhang2022bytetrack} & 63.1 & 77.3 & 80.3 & 62.0 & 61.3 & 75.2 & 77.8 & 59.6 \\
C-BIoU~\cite{yang2023hard}          & 64.1 & 79.7 & 81.1 & 63.7 & -    & -    & -    & -    \\
MotionTrack~\cite{qin2023motiontrack}& 65.1 & 80.1 & \textbf{81.1} & 65.1 & 62.8 & 76.5 & 78.0 & 61.8 \\
OC-SORT~\cite{cao2023observation}    & 63.2 & 77.5 & 78.0 & 63.4 & 62.4 & 76.3 & 75.7 & 62.5 \\
SparseTrack~\cite{liu2025sparsetrack} & 65.1 & 80.1 & 81.0 & 65.1 & 63.4 & 77.3 & \textbf{78.2} & 62.8 \\
UCMCTrack \cite{yi2024ucmctrack}         & 65.8 & 81.1 & 80.5 & 66.6 & 62.8 & 77.4 & 75.7 & 63.4 \\
\hline

\multicolumn{9}{l}{\textcolor{black}{\textbf{Motion \& Appearance:}}} \\
Quo Vadis~\cite{dendorfer2022quo}   & 63.1 & 77.7 & 80.3 & 62.1 & 61.5 & 75.7 & 77.8 & 59.9 \\
Bot-SORT~\cite{aharon2022bot}       & 65.0 & 80.2 & 80.5 & 65.5 & 63.3 & 77.5 & 77.8 & 62.9 \\
GHOST~\cite{seidenschwarz2023simple}& 62.8 & 77.1 & 78.7 & -    & 61.2 & 75.2 & 73.7 & -    \\
StrongSORT~\cite{du2023strongsort}  & 64.4 & 79.5 & 79.6 & 64.4 & 62.6 & 77.0 & 73.8 & 64.0 \\
Deep OCSORT~\cite{maggiolino2023deep}& 64.9 & 80.6 & 79.4 & 65.9 & 63.9 & 79.2 & 75.6 & 65.7 \\
DiffMOT   \cite{lv2024diffmot}                           &64.5& 79.3 &79.8 & 64.6 &
61.7 &74.9 &76.7 &60.5 \\
OFTrack \cite{song2025temporal} &64.1 &78.8 & 80.1&63.3 & 63.4&76.9 &75.6 & 62.7 \\
\textbf{GRASPTrack}          & \textbf{66.1} & \textbf{81.7} & 80.4 & \textbf{66.9} & \textbf{64.5} & \textbf{80.1} & 77.5 & \textbf{66.1} \\
\hline
\end{tabular}
\caption{Performance comparison on the MOT17 \& MOT20 test set. 
%The detection results were obtained from ByteTrack~\cite{zhang2022bytetrack}. 
The best results are shown in bold.}
\label{table1}
\end{table*}

\subsection{Depth-Aware Adaptive Noise Compensation}
The traditional KF in current MOT methods \cite{bewley2016simple, zhang2022bytetrack, cao2023observation, han2025ettrack} use a fixed process noise parameter, which limits the robustness of the tracking algorithm under occlusion and geometric ambiguity. Occluded objects may exhibit unpredictable motion patterns that are not captured by simple constant velocity models. To enhance tracking performance in such challenging conditions,  we propose the DANC, which dynamically adjusts process noise parameters.

\subsubsection{Extended State Representation.}
We extend the Kalman filter state vector to incorporate the object depth and its velocity, enabling 3D-aware motion modeling:
\begin{equation}
\mathbf{x}_t = [x, y, s, r, d, \dot{x}, \dot{y}, \dot{s}, \dot{d}]^\top,
\end{equation}
where $(x, y)$ denotes the object center in image coordinates, $s$ is the object area, $r$ is the aspect ratio, and $d$ is the estimated object depth. The terms $(\dot{x}, \dot{y}, \dot{s}, \dot{d})$ represent the respective velocities. The depth value $d$ is obtained by first employing EfficientTAM to generate precise object segmentation masks within detection bounding boxes, and then computing the average depth from the corresponding segmented regions in the depth map provided by Depth Anything v2, ensuring accurate depth representation that focuses solely on the object's actual geometry rather than background interference. The extended state representation enables a depth-aware adjustment of the Kalman noise to maintain stable predictions when the targets approach or recede rapidly.

\subsubsection{Occlusion Status Determination}
We dynamically adjust process  noise covariance based on the occlusion level of the detected object. When an object is occluded, the reliability of both its motion model and measurements decreases, increasing the uncertainty in the Kalman filter.
Let $\mathcal{D} = \{1, 2, ..., N\}$ represent all detections in the current frame, where $N$ is the total number of detections. To determine whether an object $i \in \mathcal{D}$ is occluded, the IoU is calculated between $i$ and all other objects $j \in \mathcal{D} \setminus \{i\}$. The occlusion status is determined using a depth-based criterion:
\begin{equation}
\text{occ}(i) = \begin{cases}
\text{True}, & \text{if } \exists j \in \mathcal{D} \setminus \{i\}: \text{IoU}(b_i, b_j)  \\
& \quad > \tau_{\text{IoU}} \ \text{and } d_i > d_j, \\
\text{False}, & \text{otherwise},
\end{cases}
\end{equation}
where $b_i$ and $b_j$ are the bounding boxes of objects $i$ and $j$ respectively, $d_i$ and $d_j$ are their corresponding depth values, and $\tau_{\text{IoU}}$ is the spatial overlap threshold. This process ensures that object $i$ is evaluated against all other objects in the frame to comprehensively detect the occlusion scenarios.

\subsubsection{Adaptive Noise Scaling}
For occluded objects, we adaptively scale the process noise to account for the increased uncertainty. We compute the occlusion score $\mathcal{O}_i$ as the maximum IoU overlap with all occluding objects:
\begin{equation}
\mathcal{O}_i = \max_{j \in \mathcal{D} \setminus \{i\}} \left\{ \text{IoU}(b_i, b_j) \mid \text{occ}(i) = \text{True} \right\}.
\end{equation}
The adaptive noise scaling $\lambda_i$ is then determined based on the occlusion strength:
\begin{equation}
\lambda_i = \begin{cases}
1 + \alpha \times \mathcal{O}_i, & \text{if } \text{occ}(i) = \text{True}, \\
1, & \text{otherwise},
\end{cases}
\end{equation}
%
% where $\alpha$ is a hyperparameter that controls the sensitivity to occlusion severity. Therefore, the process noise covariance is adjusted as follows:
where $\alpha$ is the occlusion sensitivity factor that controls the amplification intensity of noise scaling in response to occlusion severity. Therefore, the process noise covariance is adjusted accordingly as follows:
\begin{equation}
\mathbf{Q}_t^{\text{adaptive}} = \lambda_i \cdot \mathbf{Q}_{\text{base}},
\end{equation}
where $\mathbf{Q}_{\text{base}}$ denotes the default noise. This mechanism ensures more conservative updates in the presence of occlusions. Multiplying the process covariance by the scale factor $\lambda_i$ deliberately widens the predicted uncertainty, boosting the Kalman gain so that fresh measurements dominate whenever an object is occluded. Because $\lambda_i$ grows linearly with the occlusion score, the filter shifts smoothly from normal confidence to a more cautious mode under heavy occlusion, all without retuning the base noise matrix.

\subsection{Depth-Enhanced Observation-Centric Momentum.}
The OCM introduced in OC-SORT considers the motion direction consistency modeling of an object in the association. The original OCM calculates motion direction angles using 2D center coordinates, where the angle $\theta$  is computed as $\theta = \arctan(\frac{v_j-v_i}{u_j-u_i})$ for two points $(u_i, v_i)$ and $(u_j, v_j)$ representing object center coordinates at different time steps. Although effective in 2D scenarios, this approach cannot adequately model motion consistency when depth variations are significant. However, the OCM only relies on the velocity direction of an object in the 2D image plane and fails to capture depth-related motion consistency, particularly when objects exhibit significant displacements along the depth axis.

To address this, we propose DOCM, which operates in 3D space. Instead of computing the motion direction solely from 2D center displacements, we extend the representation to incorporate depth-aware trajectories. Let $(u_i, v_i, d_i)$ and $(u_j, v_j, d_j)$ denote the 2D center coordinates and depth values of objects at two different time steps. The corresponding 3D displacement vector $\mathbf{v}_{3D}$ is defined as:
\begin{equation}
\mathbf{v}_{3D} = [u_j - u_i,\, v_j - v_i,\, d_j - d_i]^\top.
\end{equation}

We evaluate the motion consistency by measuring the cosine similarity between the historical and current 3D motion vectors:
\begin{equation}
\mathcal{C}_{\text{VDC}} = \frac{\mathbf{v}_{\text{hist}} \cdot \mathbf{v}_{\text{curr}}}{\|\mathbf{v}_{\text{hist}}\| \cdot \|\mathbf{v}_{\text{curr}}\|},
\end{equation}
where $\mathbf{v}_{\text{hist}}$ connects two previous observations on the same trajectory and $\mathbf{v}_{\text{curr}}$ links the last track position with the current detection. 
%This formulation captures motion coherence in 3D space, making it more robust to depth variation, camera motion, and occlusions.

% dancetrack
\begin{table}[t]
\centering
\footnotesize
\resizebox{1\linewidth}{!}{
\begin{tabular}{l|p{0.75cm}p{0.75cm}p{0.75cm}p{0.75cm}}
     \hline
     Tracker   & HOTA$\uparrow$  & IDF1$\uparrow$  & MOTA$\uparrow$ & AssA$\uparrow$\\
    \hline
    
    \textbf{Motion:} & \\
    ByteTrack    & 47.3 & 52.5 & 89.5 & 31.4 \\
    C-BIoU    & 60.6 & 61.6 & 91.6	& 45.4 \\  
    MotionTrack  & 58.2 & 58.6 & 91.3 & 41.7 \\    
    OC-SORT    & 55.1 & 54.9 & 92.2 & 40.4 \\  
    SparseTrack   & 55.5 & 58.3 & 91.3 & 39.1 \\
    UCMCTrack & 63.6 & 65.0 & 88.9 & 51.3 \\

    \hline
    \textbf{Motion \& Appearance:}& \\   
    GHOST   & 56.7  & 57.7 & 91.3	& 39.8 \\
    StrongSORT     & 55.6 & 55.2 & 91.1 & 38.6 \\
    Deep OCSORT   & 61.3 & 61.5 & 92.3 & 45.8 \\
    DiffMOT        &62.3 & 63.0 & \textbf{92.8} &47.2 \\
    OFTrack & 63.4& 65.6& 91.2& 48.7\\
    \textbf{GRASPTrack} & \textbf{65.3} & \textbf{66.2} & 92.4 & \textbf{52.1} \\
  
    \hline 
\end{tabular}
}
\caption{Performance comparison on the DanceTrack test set. 
%The detection results were obtained from ByteTrack \cite{zhang2022bytetrack}. 
The best results are shown in bold.}
\label{table2}
\end{table}

\section{Experiments}

\subsection{Datasets and Evaluation Metrics}

\subsubsection{Datasets.}
We evaluate our proposed framework on three MOT benchmarks: MOT17 \cite{milan2016mot16}, MOT20 \cite{dendorfer2020mot20} and DanceTrack \cite{sun2022dancetrack}. MOT17 and MOT20 datasets are standard benchmarks commonly employed in the MOT community, featuring various challenging real-world scenarios including dense crowds, frequent occlusions, and diverse camera angles. MOT17 provides annotated pedestrian tracking data with sequences captured from different perspectives, while MOT20 presents denser scenes to evaluate tracking methods under extreme occlusion and crowd conditions. In contrast, DanceTrack specifically targets challenging tracking scenarios characterized by uniform appearances and complex, diverse motions in dance performance scenes. Utilizing these diverse benchmarks allows for a comprehensive evaluation of our framework across various and realistic tracking challenges.
\subsubsection{Evaluation.}
We adopt standard evaluation metrics commonly used in MOT, including MOTA \cite{bernardin2008evaluating}, IDF1 \cite{ristani2016performance}, HOTA \cite{luiten2021hota}, and AssA \cite{luiten2021hota}. MOTA evaluates overall tracking accuracy, combining detection accuracy with identity consistency, while IDF1 specifically measures the accuracy of maintaining object identities throughout tracking. AssA is used to evaluate the association performance. HOTA provides a balanced evaluation, capturing both association accuracy and detection performance. 

\subsubsection{Implementation Details.} 
Our proposed framework builds upon the OC-SORT baseline, integrating additional modules for depth estimation and segmentation. Specifically, we utilize the pretrained ViT-B Depth Anything v2 model \cite{yang2024depth} for zero-shot monocular depth estimation and ViT-S EfficientTAM \cite{xiong2024efficient} for precise instance segmentation. Depth maps are predicted with Depth Anything v2 and linearly scaled to the interval [0, 255]. For a fair comparison, we use the publicly available YOLOX \cite{ge2021yolox} detector weights developed by ByteTrack \cite{zhang2022bytetrack}. Because the evaluated video sequences lack camera intrinsics $(f_x, f_y)$, we first estimate them by interactively aligning a projected ground‑plane grid with each image, following the method introduced in UCMCTrack \cite{yi2024ucmctrack}. The voxel size parameter $\delta_v$ for Voxel-Based 3D IoU computation is set to 0.4, balancing computational efficiency and accuracy. For our Depth-Aware Adaptive Noise Compensation (DANC), the occlusion sensitivity factor $\alpha$ that controls the amplification intensity of noise scaling is set to 3. The spatial overlap threshold $\tau_{\text{IoU}}$, used to determine pairwise occlusion based on 3D IoU, is set to 0.6. During the association phase, we performed separate matching processes for high- and low-score detections, following the ByteTrack, with thresholds set to 0.6 and 0.1, respectively. We also employ the ReID model following the same settings as in DiffMOT \cite{lv2024diffmot}. All the experiments were conducted using a GeForce NVIDIA A100 GPU.

\subsection{Comparison with State-of-the-art Methods}
\subsubsection{MOT Challenge.}
In Table~\ref{table1}, we compare the performance of GRASPTrack with the state-of-the-art TBD methods on the MOT17 and MOT20 datasets.  To ensure fairness, all methods are evaluated using the same detection results and standardized evaluation protocols. From the comparison, our method demonstrates superior performance on both MOT17 and MOT20, achieving HOTA scores of 66.1 and 64.5, respectively. The results demonstrate the good efficiency and robustness of our method against complex scenes with occlusions.

\subsubsection{DanceTrack.} 
To demonstrate the performance of our method in complex and occluded scenarios, we test our model on the DanceTrack dataset, as shown in Table~\ref{table2}. Our results demonstrate superior performance compared to other methods and obtain a 65.3 HOTA score. The results indicate that our method can effectively handle challenging scenes with diverse motions and occlusions.

\begin{table}[t]
  \centering
  \small  % 更小字号
  \setlength{\tabcolsep}{2pt}  % 设置列之间间隔为4pt
  \begin{tabular}{c c c c | c c c}
    \hline
    Appearance & 3D IoU & DANC & DOCM & HOTA $\uparrow$ & AssA $\uparrow$ & IDF1 $\uparrow$ \\
    \hline
    & & & &52.1 & 35.3& 51.6\\
    \checkmark  &       &  &         & 58.0 & 42.3 & 57.7 \\
    \checkmark & \checkmark &         &         & 61.5 & 47.5 & 61.6 \\
    \checkmark &        & \checkmark &         & 58.9 & 42.6 & 58.6 \\
    \checkmark &  \checkmark  &  \checkmark    &  & 62.3 & 48.1 & 63.6 \\
    \checkmark &     \checkmark       & \checkmark & \checkmark &  \textbf{62.8} & \textbf{49.2} & \textbf{64.2} \\
    \hline
  \end{tabular}
  \caption{Ablation study of the GRASPTrack components. 3D IoU is Voxel-Based 3D IoU. }
  \label{tab:depth_ablation}
\end{table}

\begin{table}[t]
  \centering
\setlength{\tabcolsep}{7pt}  % 设置列之间间隔为4pt
  \small
  \begin{tabular}{c|cccc}
    \hline
VGS ($\delta_v$) & HOTA $\uparrow$ & AssA $\uparrow$ & IDF1 $\uparrow$ & FPS\\
    \hline
  0.2 & 62.3 & 48.9 & 63.8 & 9.3\\
  0.4 & \textbf{62.8} & \textbf{49.2} & \textbf{64.2} & 13.1\\
  0.6 & 62.0 & 48.8 & 63.4 & 14.0\\
   0.8 & 61.8 & 48.6 & 63.1 &14.8\\
    1.0 & 61.6 & 48.5 & 62.9 &15.1\\
    \hline
  \end{tabular}
  \caption{Impact of Voxel Grid Size (VGS) $\delta_v$ on the validation set of DanceTrack.}
  \label{tab:voxel_ablation}
\end{table}

% \begin{table}[t]
%   \centering
%   \footnotesize
%   \setlength{\tabcolsep}{8pt}  % 设置列之间间隔为4pt
%   \begin{tabular}{@{}c|ccc@{}}
%     \hline
% OSF ($\alpha$) & HOTA $\uparrow$ & AssA $\uparrow$ & IDF1 $\uparrow$ \\
%     \hline
%    0.2 &  61.9 & 48.6 & 63.7 \\
%     0.4 & 62.3 & 49.1 & 63.8 \\
%     0.6 & \textbf{62.8} & \textbf{49.2} & \textbf{64.2} \\
% 0.8 & 62.2 & 49.0 & 64.0 \\
%  1.0 & 61.8 & 48.7 & 63.5 \\
%     \hline
%   \end{tabular}
%   \caption{ Impact of Occlusion Sensitivity Factor (OSF) $\alpha$ on DanceTrack validation set.}
%   \label{tab:ocf_ablation}
% \end{table}
\begin{table}[t]
  \centering
  \footnotesize
  \setlength{\tabcolsep}{8pt}  % 设置列之间间隔为4pt
  \begin{tabular}{@{}c|ccc@{}}
    \hline
OSF ($\alpha$) & HOTA $\uparrow$ & AssA $\uparrow$ & IDF1 $\uparrow$ \\
    \hline
   1 &  61.9 & 48.6 & 63.7 \\
    2 & 62.3 & 49.1 & 63.8 \\
    3 & \textbf{62.8} & \textbf{49.2} & \textbf{64.2} \\
4 & 62.2 & 49.0 & 64.0 \\
 5 & 61.8 & 48.7 & 63.5 \\
    \hline
  \end{tabular}
  \caption{ Impact of OSF (Occlusion Sensitivity Factor) $\alpha$ on the validation set of DanceTrack.}
  \label{tab:ocf_ablation}
\end{table}

\begin{table}[t]
  \centering
    \setlength{\tabcolsep}{8pt}  % 设置列之间间隔为4pt
    \renewcommand{\arraystretch}{1.1} % 控制行高，更小值更紧凑
  \small
  \begin{tabular}{@{}c|ccc@{}}
    \hline
   Method & HOTA $\uparrow$ & AssA $\uparrow$ & IDF1 $\uparrow$ \\
    \hline
Mask-Guided Proj. &\textbf{62.8} & \textbf{49.2} & \textbf{64.2} \\
 BoundingBox Proj. & 61.7 & 48.4 & 63.1 \\
    \hline
  \end{tabular}
  \caption{Performance impact of the 3D point cloud generation strategy on the validation set of DanceTrack. }
  \label{tab:point_ablation}
\end{table}

\subsection{Ablation Study}
\label{sec:ablation}
To validate the effectiveness of our proposed depth-aware multi-object tracking framework, we conduct comprehensive ablation studies on the validation set of DanceTrack. The ablation experiments are designed to analyze four key aspects: (1) the contribution of each proposed component, (2) the impact of the Voxel Grid Size parameter, (3) the influence of the Occlusion Sensitivity Factor, and (4) the impact of the 3D Point Cloud Generation Strategy.

\subsubsection{Component Ablation}
In Table~\ref{tab:depth_ablation}, we systematically evaluated the contribution of each proposed component of GRASPTrack by progressively incorporating them into the OC-SORT baseline. The three key innovations are Voxel-Based 3D IoU, DANC and DOCM. Our experiments demonstrate that each component provides substantial improvements to baseline performance. The Voxel-Based 3D IoU computation enhances object association by replacing the traditional 2D IoU with volumetric similarity measures, enabling robust tracking in complex scenes with occlusions. The DANC improves tracking robustness by dynamically adjusting the process noise parameters based on detected occlusion events, which is particularly beneficial in occluded scenarios. The integration of DOCM provides the most substantial performance gain by extending motion consistency modeling from 2D to 3D space, effectively capturing complex motion patterns. The cumulative effect of all three components results in a comprehensive depth-aware MOT framework that significantly outperforms the baseline OC-SORT method on the DanceTrack dataset.

\subsubsection{Voxel Grid Size}
In Table \ref{tab:voxel_ablation}, we conducted extensive experiments to determine the optimal voxel grid size parameter $\delta_v$ for our Voxel-Based 3D IoU, systematically varying its value from 0.2 to 1.0 in increments of 0.2. The experimental results demonstrate that $\delta_v = 0.4$ achieves the highest tracking performance on the DanceTrack dataset, yielding the best balance among HOTA (62.8), AssA (49.2), and IDF1 (64.2) metrics. When $\delta_v$ is too small (0.2), the voxel grid becomes excessively fine-grained, leading to sparse occupancy patterns that are sensitive to depth estimation noise and resulting in increased computational overhead, as indicated by the lowest FPS (9.3). Conversely, when $\delta_v$ is too large (0.8-1.0), the voxel grid becomes overly coarse, losing critical spatial details required for accurate object discrimination, though FPS performance improves (14.8 to 15.1 FPS). The optimal value of 0.4 not only provides sufficient spatial resolution to capture meaningful volumetric overlaps and maintains robustness against depth estimation uncertainties but also achieves a reasonable computational efficiency (13.1 FPS).

\subsubsection{Occlusion Sensitivity Factor}
We investigated the impact of the occlusion sensitivity factor $\alpha$ in our depth-aware Kalman filtering mechanism by systematically varying its value from 1 to 5. As shown in Table \ref{tab:ocf_ablation}, our results reveal that $\alpha = 3$ provides the optimal balance for robust tracking performance on the DanceTrack dataset. This parameter controls the intensity of the process noise amplification during the occlusion events. When $\alpha$ is too small (1--2), the noise compensation mechanism becomes insufficient to account for the increased uncertainty during occlusion events, resulting in overconfident motion predictions that fail to adapt to unpredictable motion patterns. Conversely, when $\alpha$ is too large (4--5), the noise compensation becomes excessive, causing the Kalman filter to become overly permissive and potentially associate incorrect detections with existing tracks, leading to identity switches. The optimal value of 3 effectively addresses the motion uncertainty introduced by occlusion while maintaining sufficient discriminative power for accurate data association, and is particularly well suited for the dynamic and interactive motion patterns characteristic of group dancing scenarios.

% We investigated the impact of the occlusion sensitivity factor $\alpha$ in our depth-aware Kalman filtering mechanism by systematically varying its value from 1 to 5. As shown in Table~\ref{tab:ocf_ablation}, our results reveal that $\alpha = 3$ provides the optimal balance for robust tracking performance on the DanceTrack dataset. This parameter controls the intensity of process noise amplification during occlusion. When $\alpha$ is too small (1–2), the noise compensation becomes insufficient to account for the increased uncertainty, resulting in overconfident motion predictions. Conversely, when $\alpha$ is too large (4–5), the compensation becomes excessive, causing the Kalman filter to be overly permissive and prone to identity switches. The optimal value of 3 effectively handles motion uncertainty while maintaining sufficient discriminative power for accurate association, particularly suited to the dynamic and interactive motion patterns of group dancing scenarios.

\subsubsection{3D Point Cloud Generation Strategy}
In Table~\ref{tab:point_ablation}, we conducted experiments to validate the effectiveness of our mask-guided 3D point cloud generation strategy by comparing it with alternative approaches. We compare two different strategies: (1) Mask-guided projection using EfficientTAM to obtain segmentation masks of objects (our method) and (2) Full bounding box projection using all pixels within the detection boxes. Our experimental results on the DanceTrack dataset demonstrate that the mask-guided approach achieves the best performance with a HOTA score improvement of 1.1\% over the full bounding box method. The mask-guided strategy effectively eliminates background noise and occluder interference, leading to cleaner 3D point clouds and more accurate voxel-based 3D IoU calculations. In contrast, the full bounding box approach suffers from background contamination, particularly in crowded scenes where objects frequently overlap. Furthermore, we observe that stronger base detectors significantly enhance the effectiveness of our method. Detailed experimental results and ablation studies are provided in the Appendix.

% \section{Limitations and Future Work}

\section{Conclusion}
% This paper presents GRASPTrack, a depth-aware multi-object tracking framework that integrates monocular depth estimation with instance segmentation to achieve accurate, high-fidelity 3D point cloud reconstruction for individual objects, thus facilitating explicit 3D geometric reasoning beyond the limitations of the 2D image plane. By voxelizing these Mask-guided point clouds, our method leverages Voxel-Based 3D IoU for robust spatial association, effectively matching objects even under severe overlap and occlusion conditions. Additionally, we introduce DANC, a method that dynamically adjusts the Kalman filter’s process noise parameter according to the severity of occlusion, thereby enhancing state estimation reliability. Furthermore, we propose DOCM, extending motion consistency modeling into 3D space by incorporating depth into motion estimation, thus improving trajectory continuity. Extensive experiments validate the effectiveness and robustness of the proposed approach.

This paper presents GRASPTrack, a depth-aware multi-object tracking framework that combines monocular depth estimation and instance segmentation to reconstruct high-fidelity 3D point clouds for individual objects, enabling explicit 3D geometric reasoning beyond the 2D plane. By voxelizing these mask-guided point clouds, we compute Voxel-Based 3D IoU for robust object association under heavy occlusion. We further introduce DANC, which adaptively scales Kalman filter process noise based on occlusion severity, and DOCM, which incorporates depth into motion modeling to enhance trajectory continuity. Extensive experiments demonstrate the effectiveness and robustness of our approach in comparison to the state-of-the-art methods.

\bibliography{aaai2026}
\bibliographystyle{IEEEtran}
\end{document}